\documentclass[conference]{IEEEtran}
\usepackage{amsthm,amssymb,graphicx,multirow,amsmath,color,amsfonts}
\usepackage[update,prepend]{epstopdf}
\usepackage[noadjust]{cite}
\usepackage[latin1]{inputenc}
\usepackage{tikz} 
\usepackage{bbm} 
\usepackage{pdfpages}
\usepackage{multirow}
\usepackage{comment}

\usepackage{algorithm}
\usepackage{algpseudocode}

\usepackage{balance}

\usepackage{subfig}
\usepackage{mathrsfs}
\usepackage[normalem]{ulem}
\usepackage{booktabs}
\usepackage{optidef}
\allowdisplaybreaks 
\usepackage[colorinlistoftodos,bordercolor=orange,backgroundcolor=orange!20,linecolor=orange,textsize=scriptsize]{todonotes}

\usepackage{setspace}	

\graphicspath{{figures/}}

\newcommand{\algorihmicserver}{\textbf{Global Server}}
\algdef{SE}[SERVER]{Server}{EndServer}[1]{\algorihmicserver\ #1\ \algorihmicdo}{\algorihmicend\ \algorihmicserver}%
\algtext*{EndServer}
\newcommand{\algorihmicclient}{\textbf{Client k}}
\algdef{SE}[CLIENT]{Client}{EndClient}[1]{\algorihmicclient\ #1\ \algorihmicdo}{\algorihmicend\ \algorihmicclient}%
\algtext*{EndClient}
\makeatletter
\newcommand{\brokenline}[2][t]{\parbox[#1]{\dimexpr\linewidth-\ALG@thistlm}{\strut\raggedright #2\strut}}
\makeatother
\usepackage{float}
\usepackage{romanbar}

\makeatletter
\renewcommand{\fnum@figure}{Figure \thefigure}
\makeatother
\usepackage{tabularx}
\makeatletter
\renewcommand{\fnum@table}{Table \thetable}
\makeatother

\begin{document}
\include{notation}
\title{
FairEnergy: Contribution-Based Fairness meets Energy Efficiency in Federated Learning
}


\author{
    \IEEEauthorblockN{Ouiame Marnissi, Hajar EL Hammouti, El Houcine Bergou
        }
    \IEEEauthorblockA{College of Computing, Mohammed VI Polytechnic University, Ben Guerir, Morocco.\\
\{ouiame.marnissi, hajar.elhammouti, elhoucine.bergou\}@um6p.ma
}
}
\maketitle

\begin{abstract}

Federated learning (FL) enables collaborative model training across distributed devices while preserving data privacy.
However, balancing energy efficiency and fair participation while ensuring high model accuracy remains challenging in wireless edge systems due to heterogeneous resources, unequal client contributions, and limited communication capacity. To address these challenges, we propose FairEnergy, a fairness-aware energy minimization framework that integrates a contribution score capturing both the magnitude of updates and their compression ratio into the joint optimization of device selection, bandwidth allocation, and compression level. The resulting mixed-integer non-convex problem is solved by relaxing binary selection variables and applying Lagrangian decomposition to handle global bandwidth coupling, followed by per-device subproblem optimization.
 Experiments on non-IID data show that FairEnergy achieves higher accuracy while reducing energy consumption by up to 79\% compared to baseline strategies. 

\end{abstract}
\begin{IEEEkeywords}
Client selection, fairness, federated learning, energy efficiency, resource allocation, model compression.
\end{IEEEkeywords}

\section{Introduction}
Federated learning (FL) has emerged as a decentralized approach that allows a set of distributed devices to jointly train a shared machine learning model (ML) while keeping their data local \cite{mcmahan2017communication}. Instead of uploading raw data, devices periodically exchange model updates with a coordinating server, thus preserving privacy and efficiently exploiting the computation and storage resources available at the network edge. Despite these advantages, practical deployment in wireless and edge environments faces two major challenges: \emph{energy efficiency} and \emph{fair participation}. Communication between devices and the central server typically dominates energy consumption, while the repeated exclusion of low-resource or low-quality clients leads to severe fairness degradation and biased models. Balancing these competing objectives remains an open problem.
To reduce the communication cost in FL, various compression approaches have been introduced. They aim to decrease the size of transmitted updates by applying sparsification \cite{wangni2018gradient} or quantization\cite{reisizadeh2020fedpaq}, where only a subset or a lower-precision representation of model parameters is sent to the server. Such schemes substantially decrease communication overhead but may degrade model accuracy if not carefully designed or dynamically adapted \cite{marnissi2024adaptive}.

In parallel, resource allocation strategies have been developed to optimally distribute communication resources such as bandwidth and power among devices \cite{ nguyen2020efficient, kushwaha2025optimal}. For instance, the work in \cite{nguyen2020efficient} jointly optimize bandwidth and computation resources to minimize training costs, while authors in \cite{kushwaha2025optimal} design an energy-efficient allocation scheme for heterogeneous edge devices.

Another effective way to reduce communication overhead is client selection, where only a subset of devices participates in each round. Selecting clients with more informative or energy-efficient updates can accelerate convergence and reduce redundant communication \cite{nishio2019client, marnissi2024client, cho2022towards}. Early approaches such as FedCS \cite{nishio2019client} prioritized devices based on channel and computational conditions to meet latency constraints, while more recent studies have incorporated data-related or gradient-based criteria to improve convergence speed and model generalization \cite{marnissi2024client,cho2020client}. However, these strategies often overlook the long-term fairness of participation: devices with weaker channels or limited energy are repeatedly excluded, leading to model bias and slower convergence in heterogeneous networks.

Recent efforts have highlighted the importance of fairness in FL as clients differ in data distribution, resources, and connectivity. Naive selection can systematically exclude certain devices, leading to biased models and slower convergence. Approaches such as biased client selection analysis \cite{cho2022towards} and fairness-aware client scheduling \cite{shi2023fairness} aim to balance participation by adjusting selection probabilities or reweighting local contributions. More advanced frameworks consider multiple fairness criteria simultaneously, including participation frequency, data quality, and resource heterogeneity \cite{javaherian2024fedfair, mao2024federated}. While these methods promote balanced participation, they often ignore communication efficiency, compression, and energy constraints, which are critical in wireless and edge environments. A unified framework that jointly optimizes energy efficiency, update quality, and long-term fairness remains largely unexplored.

In this paper, we propose a Fairness-Aware Energy Minimization Framework for FL, termed \textit{FairEnergy}. 
The core idea is to promote energy-efficient and equitable participation among devices through a fairness-driven selection mechanism. Specifically, we define a score function based on the update magnitude and compression ratio. 
 These scores guide the joint optimization of device selection, compression ratio, and bandwidth allocation to minimize total communication energy while ensuring fair long-term participation across heterogeneous devices.

Our main contributions are summarized as follows:
\begin{itemize}

\item We formulate a fairness-aware energy minimization problem that jointly optimizes client selection, compression, and bandwidth allocation. The formulation integrates two important metrics: a \emph{contribution score} combining update magnitude and compression, and a \emph{long-term fairness metric} to ensure equitable participation among clients.
  \item We develop a computationally efficient solution via Lagrangian relaxation and dual decomposition. The method yields an intuitive threshold rule where a client is selected only if the benefit of its update outweighs the combined cost of energy and bandwidth. Optimal bandwidth allocation is determined via Golden Section Search (GSS), while Lagrange multipliers for bandwidth and fairness constraints are updated via subgradient ascent to ensure global feasibility.
    \item Through experiments on non-IID FMNIST data, we demonstrate that FairEnergy achieves higher model accuracy while significantly reducing total energy consumption compared to benchmark strategies. Specifically, FairEnergy requires roughly $71\%$ less energy than ScoreMax and $79\%$ less than EcoRandom to reach the same target accuracy.
\end{itemize}


The remainder of this paper is organized as follows. In section II, we present the system model. Section III introduces the fairness-aware contribution metrics. In Section IV,  we formulate the joint optimization problem and present a Lagrangian based solution in Section V. Section VI summarizes the per-round algorithm and analyzes its computational complexity. Simulation experiments are discussed in Section VII. Finally, conclusions are drawn in Section VIII.

\section{System Model}

\subsection{Learning Setup}
We consider $N$ connected clients that communicate with a central server over the wireless network. Each client trains the ML model on its private dataset $\mathcal{D}_i$ of size $|\mathcal{D}_i|$. The global learning objective is to minimize the empirical risk:
\begin{equation*}
    \min_{{\boldsymbol{w}}}
F(\boldsymbol{w})=\sum \limits_{i=1}^{N} \frac{|D_i|}{|D|}F_i(\boldsymbol{w}),
\end{equation*}
where $F_i$ is the local loss function of device $i$ and $w$ is the global model. We let the binary variable $x_i^r \in \{0,1\}$ indicate whether client $i$ is selected $(x_i^r=1)$ or not in round $r$.

The training is performed in synchronous rounds $r=1,2,\ldots$. At the beginning of round $r$, the server broadcasts the current global model $\boldsymbol{w}^{r}$ to a subset of selected devices. Each selected device then computes a local model update $\boldsymbol{u}_i^{r+1}$, i.e., the gradient of its local loss function $\nabla F_i(\boldsymbol{w}^{r+1})$ and sends it to the server. Finally, the server aggregates these local updates to obtain the new global model $\boldsymbol{w}^{r+1}$. 

In this work, we focus on the communication aspect of FL, as it remains one of the main bottlenecks in distributed training \cite{sattler2019robust}. In particular, the frequent exchange of model updates between clients and the server leads to significant communication overhead, especially in constrained networks. Consequently, we concentrate on minimizing the communication energy consumption, leaving the computation cost of local training outside the scope of the optimization \cite{wangni2018gradient}. In the following, we present the communication model adopted to characterize uplink transmission and energy consumption.

\subsection{Communication Model}
Each device $i$ communicates with the server via a wireless uplink channel. Let $B_i^r$ denote the bandwidth assigned to device $i$ in round $r$, and $\gamma_i^r \in [\gamma_{\min},1]$ denote its sparsity ratio, which represents the fraction of non-zero coefficients in the compressed model update.

The size of the transmitted payload is $\gamma_i^r S+I$, where $S$ is the size of the full model update and
$I$ is the overhead for encoding the indices of the non-zero coefficients \cite{amiri2020federated}.

We assume an additive Gaussian noise channel with channel gain $h_i^r$, and noise spectral density $N_0$. The achievable uplink rate then follows the Shannon capacity formula:
\[
R_i^r = B_i^r \log_2\!\left(1+\frac{P_ih_i^r}{N_0B_i^r}\right),
\]
where $P_i$ is the transmit power of device $i$.

The communication time of device $i$ in round $r$ is
\[
T_i^r = \frac{\gamma_i^r S+I}{R_i^r},
\]
and the corresponding communication energy is
\[
E_i^r = P_i \, T_i^r.
\]
 
A multi-objective trade-off is at the heart of efficient and fair FL. Allocating more bandwidth ($B_i^r$) or transmitting less compressed updates (higher $\gamma_i^r$) can enhance learning efficiency but escalates communication energy. Conversely, aggressive compression or selecting only clients with good channels saves immediate energy but risks discarding informative updates from other clients, thereby degrading model accuracy and more importantly violating long-term fairness among clients. Therefore, in subsequent sections, we aim to answer the following question:  How can the server jointly optimize client selection, bandwidth allocation, and compression ratios to minimize total communication energy, while simultaneously ensuring high model accuracy and guaranteeing long-term fairness for all participating clients?

\section{Fairness-Aware Contribution Metrics}

To balance efficiency with fairness in resource-constrained FL, we introduce two key metrics: a \textit{contribution score} that quantifies the utility of a compressed client update, and a \textit{long-term fairness metric} that tracks participation history.


\subsection{Contribution Score}

Client heterogeneity leads to unequal contributions to the global model. Prioritizing clients with more impactful updates can accelerate convergence \cite{marnissi2024client}, yet in resource-constrained environments, updates must often be compressed, which reduces communication cost but also limits the information available to the server. 

To capture an update's importance after compression, we define the \textit{contribution score} for client 
$i$ in round $r$ as:


\begin{equation*}
    s_i^r(\gamma_i^r) = \|u_i^r\| \cdot \gamma_i^r,
\end{equation*}
where $\|u_i^r\|$ is the $L_2$ norm of the update $u_i^r$. The $L_2$ norm of the update reflects the overall magnitude of the changes contributed by the client, while the compression factor scales this magnitude according to the information actually sent. This score provides a simple and practical measure of each client's effective contribution.

\subsection{Long-Term Fairness Metric}
The proposed fairness metric is framed as long-term participation fairness, in that we track the effective participation rate of each client and enforce a minimum threshold on that rate. Prior works have argued for fairness in client selection by ensuring each client gets an equitable number of participation opportunities across rounds \cite{javaherian2024fedfair} and by incorporating clients' past selection history in the selection process \cite{shi2023fairness}.

To capture this long-term participation, we adopt an exponential moving average (EMA):
\begin{equation} \label{fairness_def}
 q_i^r = \rho\, q_i^{r-1} + (1-\rho)\, x_i^r,   
\end{equation}
where $\rho$ controls the memory of past rounds and the initialization of $q_i^{0}$ is discussed in Section VI-A.  
This formulation allows us to maintain a smooth and responsive measure of participation: clients that have been selected less frequently in recent rounds have a lower $q_i^r$ and are thus more likely to be prioritized in future selections. By tuning $\rho$, we can control how strongly recent participation affects the metric, providing a flexible way to ensure long-term fairness. This also enables the selection framework to dynamically boost under-selected clients and maintain equitable participation over time. Finally, we impose a minimum participation threshold:
\[
q_i^r \geq \pi_{\min}, \quad \forall i \in \{1,\ldots,N\},
\]
ensuring that every client contributes to the learning process at least at a minimal rate $\pi_{\min}$ over time.

By combining ${s_i}^r$ and $q_i^r$, the framework simultaneously prioritizes high-quality updates and guarantees fair long-term participation across all devices.
These metrics serve as the basis for jointly selecting devices, allocating resources, and adapting compression ratios.

In the next section, we formalize this design as a fairness-aware energy minimization problem, where the goal is to optimize communication energy while ensuring both update quality and equitable participation.
\section{Problem Formulation}

In this section, we formulate the fairness-aware energy minimization problem. 
At each global round $r$, the server aims to select a subset of participating devices while jointly optimizing 
their communication parameters (compression ratio and bandwidth allocation) to minimize the total communication energy.
Each device $i$  contributes to the global model with a score $s_i^r(\gamma_i^r)$ that reflects the relevance and quality of its local update. To prevent over-representation of a few high-scoring devices, we introduce a long-term fairness constraint ensuring that each device participates in the training process with at least a minimum rate $\pi_{min}$. This translates into the following problem:
\begin{mini!}
{\{x_i^r,\,\gamma_i^r,\,B_i^r\}_{i=1}^N}
{\sum_{i=1}^N \Big( x_i^r E_i^r(\gamma_i^r,B_i^r) \! -\! \eta\, x_i^r{s}_i^r(\gamma_i^r) \Big) \label{mainobj}}
{\label{GeneralOptimizationPb}}{}
\addConstraint{\sum_{i=1}^N x_i^r B_i^r \le B_{\mathrm{tot}} \label{band} }
{}{}
\addConstraint{\gamma_{\min} \le \gamma_i^r \le 1, \quad\forall i \label{gamma}}
{}{}
\addConstraint{x_i^r \in \{0,1\}, \quad\forall i \label{user_selection}}
{}{}
\addConstraint{q_i^r \;\ge\; \pi_{min}, \quad\forall i \label{fairnes}.} 
\end{mini!}
Constraint (\ref{band}) ensures that the total allocated bandwidth does not exceed the available system budget. Constraint (\ref{gamma}) bounds the compression ratio within a feasible range. The binary variable $x_i^r$ in (\ref{user_selection}) indicates whether device $i$ participate in round $r$. Finally, constraint (\ref{fairnes}) ensures that each device maintains a minimum participation ratio over time, where $q_i^r$ is the EMA fairness metric defined in equation (\ref{fairness_def})\footnote{Although the global loss convergence is not explicitly added as a constraint in the studied optimization, our design is supported by our prior theoretical work \cite{marnissi2024client}, which established that selection based on update magnitude accelerates convergence. Here, we extend this result to a more practical setting with compression and fairness criteria, and empirically show convergence in Section~VII, leaving a full theoretical analysis for future work.}.

Problem~(\ref{GeneralOptimizationPb}) is mixed-integer, non-convex optimization problem. The binary participation variables $x_i^r$ create combinatorial complexity, while the coupling between selection, compression, and bandwidth makes joint optimization challenging. 
To address these challenges, we adopt a Lagrangian relaxation approach \cite{bertsekas1997nonlinear} that introduces dual variables associated with bandwidth and fairness constraints. The problem is decomposed to simpler per-device subproblems. Each device can then optimize its local compression and bandwidth decisions independently, while the server iteratively adjusts dual variables to enforce global fairness and bandwidth feasibility.

In the following section, we detail the Lagrangian formulation and the iterative resolution process adopted in our setup.
\section{Problem Resolution with Lagrangian Relaxation}

To efficiently solve the per-round optimization problem (\ref{GeneralOptimizationPb}), we adopt a relaxation-decomposition strategy that enables distributed per-device updates while ensuring both bandwidth feasibility and long-term fairness. Concretely, we first relax the binary selection variables $x_i^r \in \{0,1\}$ to the continuous interval $[0,1]$. Next, we form the partial Lagrangian by dualizing only the coupling constraints (bandwidth and fairness), which allows the problem to decompose across devices and enables efficient distributed optimization. Finally we recover feasible binary decisions through a repair step.

\subsection{Lagrangian Relaxation and Problem Decomposition}

After relaxing $x_i^r \in \{0,1\}$ to $x_i^r \in [0,1]$, we introduce dual variables $\lambda^r \ge 0$ for the bandwidth constraint in (\ref{band}), and $\mu_i^r \ge 0$ for the fairness constraint in (\ref{fairnes}). The partial Lagrangian is therefore

\begin{equation}
\begin{aligned}
\mathcal{L}\!(\mathbf{x}^r,\boldsymbol{\gamma}^r,\mathbf{B}^r;\lambda^r,\boldsymbol{\mu^r})
= \sum_{i=1}^N  x_i^r(E_i^r(\gamma_i^r,B_i^r) - \eta\,{s}_i^r(\gamma_i^r)) \\
\quad + \lambda^r\Big(\sum_{i=1}^N x_i^r B_i^r - B_{\mathrm{tot}}\Big)
+ \sum_{i=1}^N \mu_i^r (\pi_{min} - q_i^r).
\end{aligned}
\end{equation}

By substituting the explicit form of the fairness metric in equation ($\ref{fairness_def}$) into the Lagrangian and rearranging the terms, the global Lagrangian can be separated into a sum of independent, per-device Lagrangians:

\begin{equation}
\begin{split}
\mathcal{L}_i(\!x_i^r, \gamma_i^r\!, \!B_i^r;\!\lambda^r\!\!,\boldsymbol{\mu^r}\!) \!\!= \!x_i^r \!\bigg(\!\! & E_i^r\!(\gamma_i^r, B_i^r)\!\! + \!\!\lambda^r \!B_i^r \!\! - \!\eta s_i^r\!(\!\gamma_i^r\!) \!- \!\mu_i^r\! (1\!-\!\rho) \!\!\bigg) \\
& + \mu_i^r \left( \pi_{\min} - \rho q_i^{r-1}\! \right),
\end{split}
\end{equation}
where the term $\mu_i^r \left( \pi_{\min} - \rho q_i^{r-1} \right)$ is a constant with respect to the current round's decision variables.

Thus, for fixed duals \((\lambda^r, \boldsymbol{\mu}^r)\), the global problem decomposes into $N$ independent local minimization problems. 
\[
\min_{x_i^r\in[0,1],\ \gamma_i^r\in[\gamma_{\min},1],\ B_i\ge0}
\; \mathcal{L}_i(x_i^r,\gamma_i^r,B_i^r).
\]
\subsection{Per-Device Minimization}

Because the function \(\mathcal{L}_i\) is \emph{affine} in \(x_i^r\), the minimizer over $x_i^r \in [0,1]$ lies at the boundaries of the feasible interval.

\[
x_i^{r} =
\begin{cases}
1, & E_i^r(\gamma_i^r,B_i^r) + \lambda^r B_i^r < \eta\,{s}_i^r(\gamma_i^r) + \mu_i^r(1-\rho),\\
0, & \text{otherwise}.
\end{cases}
\]

Hence, the relaxed solution yields a binary decision:  the device participates ($x_i=1$) only if the benefit of its update outweighs the associated energy and bandwidth costs.
We therefore consider the two possible cases:

\noindent \textit{Case 1: Device not selected} (\(x_i^r = 0\))\\
The Lagrangian reduces to a constant:
\[
\mathcal{L}_i^{\mathrm{not}} = \mu_i^r(\pi_{\min} - \rho q_i^{r-1}),
\]
and no further optimization over \(\gamma_i^r\) or \(B_i^r\) is required.

\noindent \textit{Case 2: Device selected} (\(x_i^r = 1\))\\
We must solve the joint compression and bandwidth allocation subproblem:
\[
\mathcal{L}_i^{\mathrm{sel}} = \min_{\gamma_i^r, B_i^r} \phi_i^r(\gamma_i^r, B_i^r) - \mu_i^r(1-\rho) + \mu_i^r(\pi_{\min} - \rho q_i^{r-1}),
\]
where the core optimization is captured in the function:
\begin{equation} \label{eq:phi_def}
\phi_i^r(\gamma_i^r, B_i^r) = E_i^r(\gamma_i^r, B_i^r) + \lambda^r B_i^r - \eta s_i^r(\gamma_i^r).
\end{equation}
Only this term depends on the decision variables \(\gamma_i^r\) and \(B_i^r\).




\subsection{Joint Optimization of $(\gamma_i^r,B_i^r)$ for Selected Clients}
When a client is selected, the minimization of the Lagrangian over $(\gamma_i^r,B_i^r)$  reduces to the following problem 
\[
\min_{\gamma_i^r,B_i^r} \phi_i^r(\gamma_i^r,B_i^r).
\] 

The function \(\phi_i^r\) is not convex in \(B_i^r\), but exhibits a \emph{unimodal} shape:  
for small \(B_i^r\), transmission energy dominates; as \(B_i^r\) increases, energy decreases sharply, then flattens as the achievable rate saturates; finally, the linear term \(\lambda^r B_i^r\) grows, increasing the total cost.  
This decreasing-flattening-increasing pattern produces a single minimum, making the problem well-suited to derivative-free one-dimensional search methods such as the Golden Section Search (GSS) \cite{wang2014optimization,kiefer1953sequential}.

We therefore solve the joint minimization in three steps:

\begin{enumerate}
    \item Discretize $\gamma_i^r$ on a small finite grid $\Gamma = \{\gamma^{(1)}, \dots, \gamma^{(G)}\}$.
    \item  For each $\gamma^{(g)} \in \Gamma$, find $B_i^{r,\star}(\gamma^{(g)}) = \arg\min_{B_i^r} \phi_i^r(\gamma^{(g)},B_i^r)$ using GSS.
    \item   
    Evaluate $\phi_i^r(\gamma^{(g)},B_i^{r,\star}(\gamma^{(g)}))$ and choose  
    $(\gamma_i^{r,\star},B_i^{r,\star}) = \arg\min_{\gamma^{(g)}\in\Gamma}
\phi_i^r(\gamma^{(g)},B_i^{r,\star}(\gamma^{(g)}))$.
\end{enumerate}

\section{Algorithm and Dual Updates}
In the following, we provide a brief summary of the per- resolution of the FairEnergy optimization algorithm, followed by an analysis of its computational complexity.
\begin{algorithm}[H]
\caption{Per-Round FairEnergy Optimization} \label{algo:resolution}
\begin{algorithmic}[1]
\State\textbf{Input:} previous fairness metrics $q_i^{r-1}$, parameters $\eta, \rho, \Gamma$, step sizes $\alpha_\lambda$, $\alpha_\mu$ $> 0$
\State \textbf{Initialize:} dual variables $\lambda^r, \boldsymbol{\mu}^r$
\Repeat
    \For{each device $i=1,\dots,N$ }
        \State For each $\gamma_i^r \in \Gamma$, use GSS to find $B_i^{r,\star}(\gamma_i^r)$ that minimizes $\phi_i^r$ defined in equation (\ref{eq:phi_def}).
        \State Select $(\gamma_i^{r,\star},B_i^{r,\star})=\arg\min_{\gamma_i^r \in \Gamma} \phi_i^r(\gamma_i^r,B^{r,\star}(\gamma_i^r))$.
        \State Determine  selection:
        \[
        x_i^{r,*} \!\!=\!
        \begin{cases}\!
        1, &\!\!\!\! E_i^r(\gamma_i^{r,\star}\!,B_i^{r,\star}) \!+\! \lambda^r \!B_i^{r,\star}\! \!\!< \!\eta\,{s}_i^r(\gamma_i^{r,\star})\! + \!\mu_i^r(1\!-\!\rho),\\
       \! 0, &\!\!\!\! \text{otherwise.}
        \end{cases}
        \]
        \State Update fairness metric: $    q_i^{r+1} = \rho q_i^r + (1-\rho) x_i^{r,\star}$
        \State Update fairness dual:
        \[
        \mu_i^r \leftarrow \big[\mu_i^r + \alpha_\mu(\pi_{min} - \rho q_i^{r-1} -(1-\rho)x_i^{r, \star})\big]^+
         \]
    \EndFor
    
    \State Server updates bandwidth dual: 
    \[
    \lambda^r \leftarrow \big[\lambda^r + \alpha_\lambda(\sum_i x_i^{\mathrm{r,\star}} B_i^{r,\star} - B_{\mathrm{tot}})\big]^+
    \]
\Until{convergence}
\end{algorithmic}
\end{algorithm}

\subsection{Algorithm}
Algorithm \ref{algo:resolution} summarizes the per-round resolution of the proposed FairEnergy optimization. For a given round $r$, each device $i$ first solves a local optimization over the feasible compression ratios $\gamma_i^r\in\Gamma$ to obtain the corresponding optimal bandwidth allocation $B_i^{r,\star}(\gamma_i^r)$ via GSS. The fairness metric $q_i^{r}$ is subsequently updated using an EMA to capture the device's participation history. \textit{For feasibility, we initialize $q_i^0$ sufficiently large so that in
the early rounds, fairness is not restrictive and selection is
mainly driven by minimizing energy and maximizing score.
As rounds progress, $q_i^r$ naturally evolves to enforce fairness.} Finally, the dual variables $\lambda$ and $\boldsymbol{\mu}$ associated respectively with the total bandwidth and fairness constraints are updated as shown in lines $10$ and $12$ of Algorithm \ref{algo:resolution}. These updates follow the standard projected subgradient ascent rule commonly used in dual decomposition methods \cite{bertsekas1997nonlinear, palomar2006tutorial}, where the subgradients correspond to the constraint violations of the bandwidth and fairness limits.
In practice, these dual variables are typically refined over several inner iterations to ensure convergence of the Lagrangian subproblem within each communication round. However, to keep notation concise and the algorithm readable, we present only one generic update per round, which implicitly captures this iterative process. 

\subsection{Complexity Analysis}

The computational complexity of FairEnergy is dominated by the per-device joint optimization of the compression ratio $(\gamma_i^r$ and bandwidth $B_i^r$. For each device, $\gamma_i^r$ is discretized over a grid of size $G$, and for each grid point, $B_i^r$ is optimized via the GSS method with $T_{\mathrm{GSS}}$ function evaluations, yielding a per-device complexity of $\mathcal{O}( G\, T_{\mathrm{GSS}})$. Subsequent computations of the selection metric, relaxed selection, and local fairness updates are negligible in comparison. The global bandwidth dual update $\lambda^r$ requires a sum over $N$ devices, giving $\mathcal{O}(N)$ complexity. Consequently, the total per-round complexity is $\mathcal{O}(N \, G \, T_{\mathrm{GSS}})$.


\section{Simulation Results}

In this section, we evaluate the performance of our proposed framework in terms of test accuracy and communication energy.  Experiments are implemented in Keras with TensorFlow. 
We train a convolutional neural network with approximately $2$ million parameters on a non-IID partitioned FMNIST, a dataset of fashion products from $10$ categories. 
We consider a network with $N=50$ devices. The total available bandwidth is $B_{\mathrm{tot}} = 10$ MHz, and transmission power for each device is uniformly distributed in $[0.1,0.3]$ mW. 
To model heterogeneous data distributions, local datasets are sampled using a Dirichlet distribution $Dir_K(\beta)$ with concentration parameter $\beta=0.3$ \cite{li2022federated}. 
Compression ratios are constrained in $[0.1, 1]$, and long-term fairness is enforced with threshold $\pi_{\min} = 0.2$. The participation memory for fairness tracking is $\rho = 0.6$ and the learning rate is of $0.01$. 

To ensure a fair comparison, the number of selected devices in the baseline schemes is fixed to the mean number of devices selected by FairEnergy across rounds.

We evaluate the performance of FairEnergy to the following baselines designed to isolate components of our approach:

\begin{itemize}
\item ScoreMax: selects devices with the highest contribution scores, ignoring energy and fairness. To isolate the effect of contribution score, updates are transmitted at full precision ($\gamma_i = 1$), and the available bandwidth $B_\text{tot}$ is equally divided among the selected devices. This approach isolates the effect of importance-driven selection and has been extensively investigated in \cite{marnissi2024client,chen2020optimal}.
    \item EcoRandom: a communication-efficient random selection baseline that removes both fairness and contribution-awareness. In each round, devices are chosen randomly, and all selected devices transmit using the minimum compression ratio and bandwidth observed in FairEnergy. This configuration drives EcoRandom toward the lowest possible communication energy, in line with the principle of minimizing transmission cost explored in several state-of-the-art energy-aware FL studies \cite{marnissi2024adaptive,kim2023green}.

\end{itemize}

These benchmarks allow us to assess the trade-offs between energy efficiency, model performance, and fairness under different device selection policies.

\begin{figure*}[t]
\minipage{0.33\textwidth}
  \includegraphics[width=\linewidth]{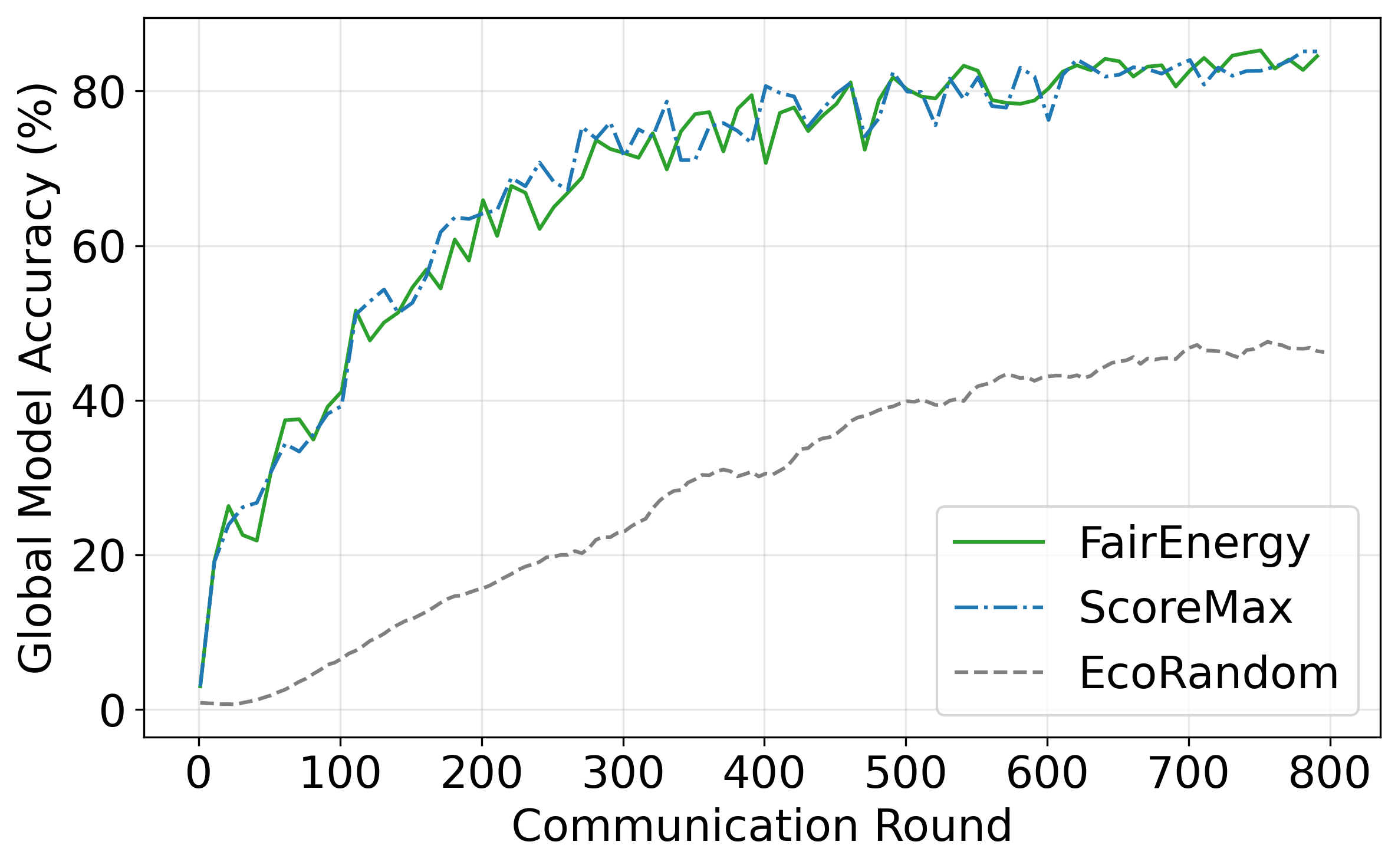}
  \caption{Test accuracy per round.}\label{fig:acc}
\endminipage\hfill
\minipage{0.33\textwidth}
  \includegraphics[width=\linewidth]{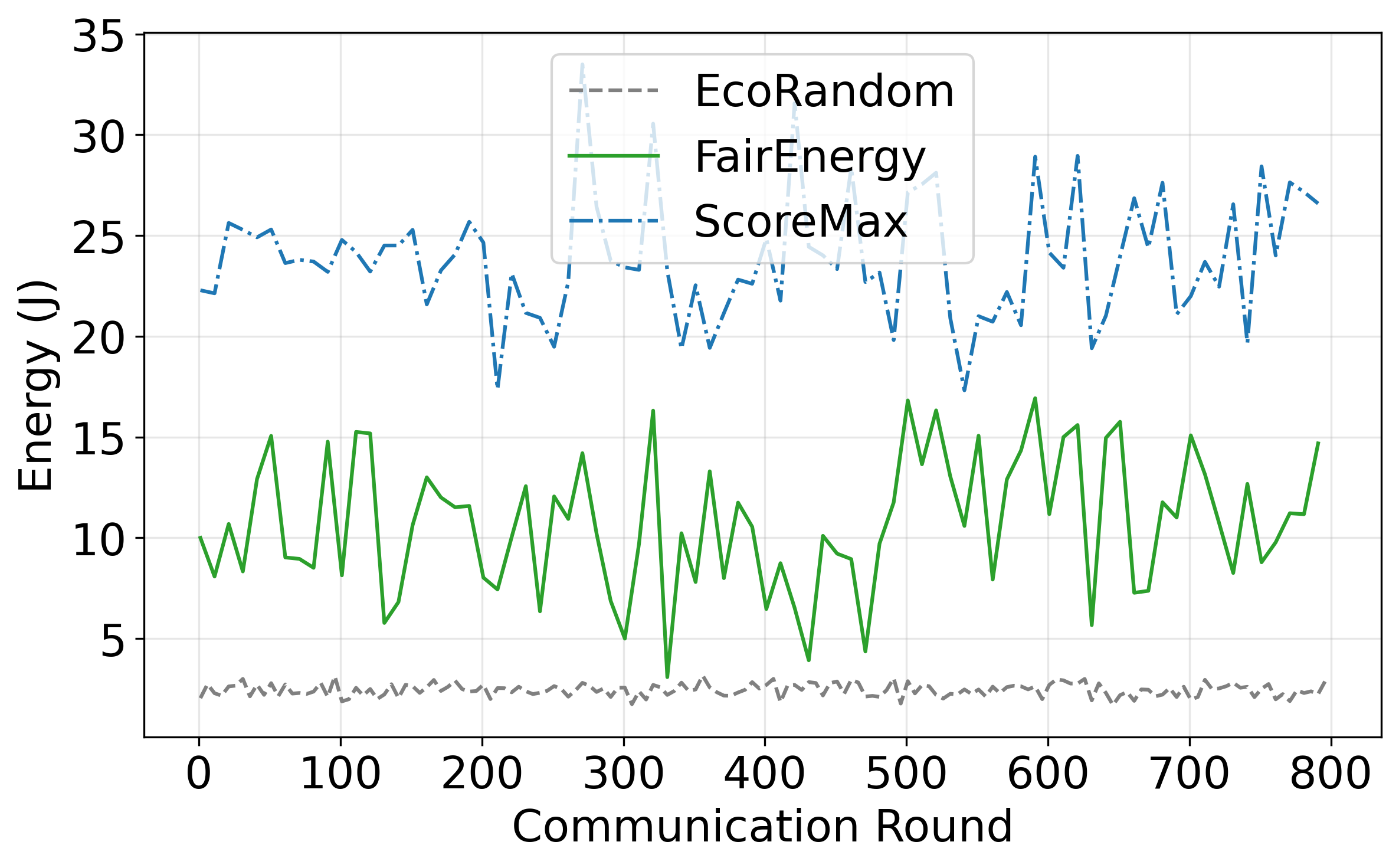}
  \caption{Energy consumption per round.}\label{fig:energy}
\endminipage\hfill
\minipage{0.33\textwidth}%
  \includegraphics[width=\linewidth]{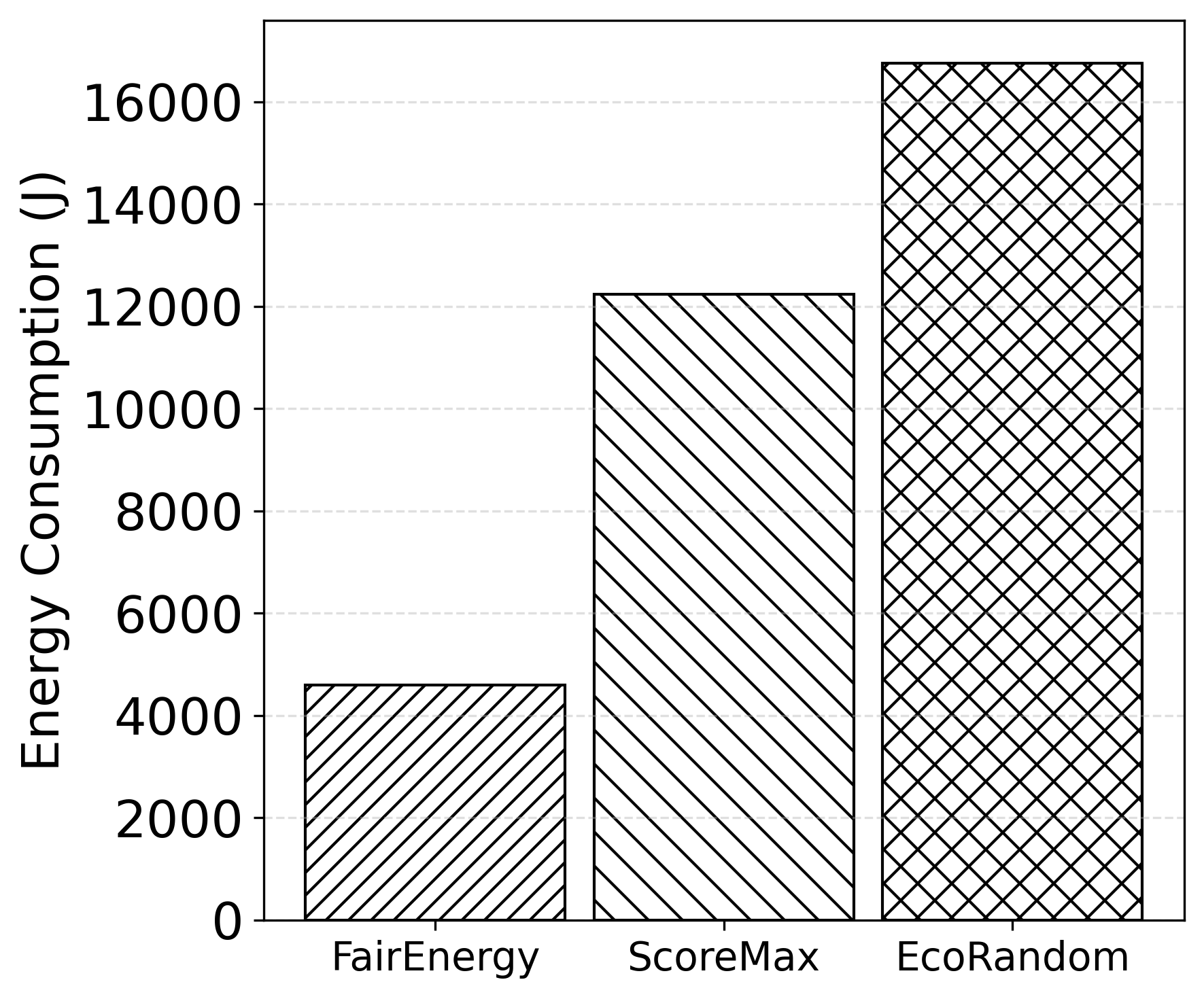}
  \caption{Total energy needed to achieve a target accuracy.}\label{fig:total_energy}
\endminipage
\end{figure*}

Figure \ref{fig:acc} shows that FairEnergy and ScoreMax achieve the highest test accuracy. Both methods select devices based on their contribution importance, leading to the inclusion of clients providing the most informative updates. 
In contrast, EcoRandom has slow convergence over the rounds mainly because of the large compression levels applied to reduce transmission energy. Notably, FairEnergy also applies compression, yet it maintains accuracy comparable to ScoreMax throughout the rounds. This is because FairEnergy balances contribution with fairness in client participation, ensuring that both high-quality and underrepresented devices contribute to model training.

In Figure \ref{fig:energy}, we report the average energy consumption per global round. As expected, EcoRandom achieves the lowest per-round energy consumption due to its aggressive compression and favorable bandwidth allocation. FairEnergy follows closely, consuming slightly more energy. In contrast, ScoreMax operates with no compression and with uniform bandwidth across devices, which leads to consistently higher energy consumption over the rounds.

In Figure \ref{fig:total_energy}, we compare the total cumulative energy required to reach a target accuracy of $80\%$. 
The results clearly highlights the effectiveness of the proposed FairEnergy framework, which achieves the target accuracy while consuming substantially less energy than the baselines. Specifically, FairEnergy requires roughly $71\%$ less energy than ScoreMax and $79\%$ less than EcoRandom. While EcoRandom consumes less energy per round, its slower convergence leads to a higher overall energy cost to reach the desired accuracy. ScoreMax, which operates without compression and uniform bandwidth, also requires significantly more energy due to inefficient communication. These results demonstrate that combining contribution-aware device selection with fairness-driven participation effectively reduces the total energy needed to achieve a target accuracy.
 
\begin{table}[H]
\centering
\caption{Device participation statistics across strategies.}
\begin{tabular}{lccc}
\toprule
\textbf{Strategy} & \textbf{Min} & \textbf{Max} & \textbf{Std} \\
\midrule
FairEnergy & 401 & 413 & 2.85 \\
ScoreMax   & 50  & 697 & 180.93 \\
EcoRandom  & 369 & 423 & 15.79 \\
\bottomrule
\end{tabular}
\label{tab:participation_stats}
\end{table}
Table \ref{tab:participation_stats} highlights how FairEnergy enforces fairness in device participation, maintaining a tight selection range across devices. EcoRandom, being a random selection baseline, shows moderate variability, while ScoreMax prioritizes only the most informative clients, resulting in highly uneven participation.

\section{Conclusion}
In this work, we proposed FairEnergy, a joint fairness and energy-aware framework for FL that optimizes device selection, bandwidth allocation, and compression ratios. By using a contribution score combining update magnitude and compression ratio, along with long-term fairness constraints, our approach ensures efficient communication, equitable participation, and faster model convergence . The proposed per-round resolution, based on Lagrangian relaxation and efficient per-device minimization, enables scalable implementation across large networks. Experiments demonstrate that FairEnergy achieves substantial energy savings while maintaining high model accuracy. Future work will extend this framework to dynamic channel environments and joint optimization of computation and communication resources.

 \balance
\bibliographystyle{IEEEbib}
\bibliography{references}
\end{document}